%% file: main.tex
\documentclass[10pt,twocolumn,letterpaper]{article}

\usepackage[pagenumbers]{wacv}

\usepackage{graphicx}
\usepackage{multirow}
\usepackage{adjustbox}
\usepackage{amsmath}
\usepackage{amssymb}
\usepackage{booktabs}
\usepackage{makecell}

\usepackage{algorithm}
\usepackage{algorithmicx}
\usepackage{algpseudocode}
\usepackage{url}
\usepackage{fontawesome5}

\newcommand{\MissionGNN}{\textsc{MissionGNN} }

\usepackage[pagebackref,breaklinks,colorlinks]{hyperref}

\usepackage[capitalize]{cleveref}
\crefname{section}{Sec.}{Secs.}
\Crefname{section}{Section}{Sections}
\Crefname{table}{Table}{Tables}
\crefname{table}{Tab.}{Tabs.}

\begin{document}

\title{\textsc{MissionGNN}: Hierarchical Multimodal GNN-based Weakly Supervised Video Anomaly Recognition with Mission-Specific Knowledge Graph Generation}

\author{Sanggeon Yun\\
University of California, Irvine\\
Irvine, CA, 92617\\
{\tt\small sanggeoy@uci.edu}
\and
Ryozo Masukawa\\
University of California, Irvine\\
Irvine, CA, 92617\\
{\tt\small rmasukaw@uci.edu}
\and
Minhyoung Na\\
Kookmin University\\
Seoul, South Korea, 02707\\
{\tt\small minhyoung0724@kookmin.ac.kr}
\and
Mohsen Imani\thanks{Corresponding Author}\\
University of California, Irvine\\
Irvine, CA, 92617\\
{\tt\small m.imani@uci.edu}
}
\maketitle

%%%%%%%%% ABSTRACT
\begin{abstract}
   \input{Sections/0_Abstract}
\end{abstract}

%%%%%%%%% BODY TEXT
\section{Introduction}
\label{sec:intro}

\input{Sections/1_Introduction}

\section{Background and Related Works}

\input{Sections/2_RelatedWorks}

\section{Methodology}

\input{Sections/3_Methodology}

\section{Experiments}

\input{Sections/4_Experiments}

\section{Discussion}
\input{Sections/5_Discussion}

\section{Conclusions}
\input{Sections/6_Conclusions}

\section*{Acknowledgements}
This work was supported in part by the DARPA Young Faculty Award, the National Science Foundation (NSF) under Grants \#2127780, \#2319198, \#2321840, \#2312517, and \#2235472, the Semiconductor Research Corporation (SRC), the Office of Naval Research through the Young Investigator Program Award, and Grants \#N00014-21-1-2225 and \#N00014-22-1-2067, and Army Research Office Grant \#W911NF2410360. Additionally, support was provided by the Air Force Office of Scientific Research under Award \#FA9550-22-1-0253, along with generous gifts from Xilinx and Cisco.

%%%%%%%%% REFERENCES
{\small
\bibliographystyle{ieee_fullname}
\bibliography{egbib}
}

\newpage
\section*{Appendix}
\input{Appendices/appendix}

\end{document}

%% file: Sections/0_Abstract.tex
In the context of escalating safety concerns across various domains, the tasks of Video Anomaly Detection (VAD) and Video Anomaly Recognition (VAR) have emerged as critically important for applications in intelligent surveillance, evidence investigation, violence alerting, etc. These tasks, aimed at identifying and classifying deviations from normal behavior in video data, face significant challenges due to the rarity of anomalies which leads to extremely imbalanced data and the impracticality of extensive frame-level data annotation for supervised learning. This paper introduces a novel hierarchical graph neural network (GNN) based model \MissionGNN that addresses these challenges by leveraging a state-of-the-art large language model and a comprehensive knowledge graph for efficient weakly supervised learning in VAR. Our approach circumvents the limitations of previous methods by avoiding heavy gradient computations on large multimodal models and enabling fully frame-level training without fixed video segmentation. Utilizing automated, mission-specific knowledge graph generation, our model provides a practical and efficient solution for real-time video analysis without the constraints of previous segmentation-based or multimodal approaches. Experimental validation on benchmark datasets demonstrates our model's performance in VAD and VAR, highlighting its potential to redefine the landscape of anomaly detection and recognition in video surveillance systems. The code is available here: \faIcon{github}\ \url{https://github.com/c0510gy/MissionGNN}.

%% file: Sections/1_Introduction.tex
The escalating concern for safety across various sectors has heightened the importance of tasks like Video Anomaly Detection (VAD) and Video Anomaly Recognition (VAR). These tasks are crucial for applications such as intelligent surveillance~\cite{bao2022hierarchical, feng2021convolutional}, evidence investigation~\cite{nanda2022soft}, and violence alerting~\cite{sahay2022real, islam2023iot}, where the goal is to automatically identify and classify activities that deviate from normal patterns in video data~\cite{suarez2020survey}. VAD and VAR can potentially be approached through supervised learning to detect anomalies~\cite{bai2019traffic, wang2019anomaly}; however, the rarity of such anomalies leads to highly imbalanced datasets, complicating the training process. Moreover, the extensive effort and cost required for frame-level video data annotation render supervised learning with frame-level annotations impractical. Consequently, many researchers have shifted their focus towards weakly-supervised learning~\cite{li2022scale, sultani2018real, tian2021weakly, wu2021learning, zanella2023delving, zhou2023batchnorm}, which relies on video-level annotations to train models. This shift addresses the challenges posed by data imbalance and annotation costs but introduces new complexities in achieving accurate frame-level anomaly detection and recognition.

Recent advancements in tackling these challenges involve the use of pre-trained models to leverage the rich, embedded information they offer, thereby enhancing the distinction between normal and abnormal frames~\cite{zanella2023delving, zhou2023batchnorm, sultani2018real, wang2023actionclip, pu2023learning}. However, for VAR, the intricacies go beyond what pre-trained vision models can discern, necessitating finer granularity. To address this, some studies~\cite{zanella2023delving, wang2023actionclip, wang2021actionclip, xu2021videoclip, pu2023learning} have explored large multimodal neural networks~\cite{radford2021learning, singh2022flava} capable of extracting more nuanced distributions, employing methods like learnable prompts~\cite{zhou2022learning} or adapters~\cite{wang2023actionclip} to refine model differentiation. This approach, while innovative, demands substantial computational resources due to the gradient flow through these large models~\cite{ha2023meta}. Furthermore, it is incompatible with the prevailing trend towards large pre-trained models, which are typically accessed via APIs, making it impossible to compute gradients directly.

Moreover, the application of multiple-instance learning (MIL) with fixed video segmentation has been a popular simplified strategy among researchers~\cite{sultani2018real}. By dividing videos into fixed-size or fixed-number segments and introducing a transformer model to capture not only short-term relationships but also long-term relationships between segments, these models have shown improved performance~\cite{zanella2023delving, wang2023actionclip, zhong2019graph, tian2021weakly, li2022scale, wu2021learning}. Yet, this method struggles with the variability of anomaly event durations e.g., events that happen in very short periods or video length is much longer than the event length, and the practical limitations of real-time analysis, where future frame data is unavailable, and past frame data is potentially infinite. Such reliance on long-term relationship modeling further complicates the deployment of these models in real-world scenarios, where rapid and continuous analysis is essential.

This paper proposes a novel approach \MissionGNN to tackle VAR by introducing a novel hierarchical graph neural network (GNN) based model that avoids the computational burdens associated with gradient computation of large multimodal models and introduces a fully frame-level training methodology without using MIL. Unlike prior work that limitedly utilized GNNs for temporal information processing~\cite{zhong2019graph, ullah2023ad}, our model uses GNN to capture semantic information using a mission-specific knowledge graph (KG) that is automatically generated by our KG generation framework using a large language model namely GPT-4~\cite{achiam2023gpt}, and a large knowledge graph namely ConceptNet~\cite{speer2017conceptnet}. Instead of flowing gradient to a large multimodal model, our approach utilizes a built mission-specific knowledge graph and only uses the large multimodal model without any gradient computation to have initial embeddings for each node that will be used in our GNN. 
After that, our approach uses a small transformer to capture only short-term relationships. This innovative strategy enables effective anomaly detection without the need for gradient computations on large models, relying instead on light models while offering a practical, efficient, and scalable solution to the challenges of weakly-supervised learning in video anomaly recognition.

In summary, our work represents a fundamentally novel contribution to the field, offering the following key advancements:

\begin{itemize}
    \item To the best of our knowledge, we propose \MissionGNN a novel hierarchical GNN-based model with an automated mission-specific knowledge graph generation framework to have an efficient weakly supervised video anomaly recognition model. Through our proposed mission-specific knowledge graph generation framework, our approach can be adopted in various other tasks.
    \item Unlike previous works on using pre-trained large multimodal models, our approach does not require heavy gradient computation on the pre-trained large multimodal models providing efficient training. At the same time, our approach provides additional flexibility that also can be implemented using large pre-trained models that are exclusively accessible through APIs which we do not have access to the weights of the model.
    \item Inspired by human intuition, we introduce completely frame-level training and inference that does not require video segmentation which can affect the model's performance by merging multiple frames into fixed-size segments. Furthermore, this makes our model more practical in real applications by not using future segments or past segments suitable for real-time applications.
\end{itemize}

%% file: Sections/2_RelatedWorks.tex
\subsection{Weakly Supervised Video Anomaly Detection and Recognition}

Video anomaly detection has traditionally relied on fully supervised methods requiring frame-level annotations, which are costly and time-consuming~\cite{bai2019traffic, wang2019anomaly}. To address these challenges, weakly supervised approaches using video-level labels~\cite{li2022scale, sultani2018real, tian2021weakly, wu2021learning}, one-class classification~\cite{liu2021hybrid, lv2021learning}, and unsupervised methods~\cite{narasimhan2018dynamic, zaheer2022generative} have gained traction. Weakly supervised techniques, in particular, offer a balance between performance and annotation costs, making them popular in recent research.

MIL~\cite{sultani2018real}, which divides videos into segments and labels them uniformly, is a key method in weakly supervised video anomaly detection. Enhancements like internal loss constraints~\cite{zhang2019temporal} and pseudo labeling~\cite{zhong2019graph} have enabled models to capture both short- and long-term relationships~\cite{tian2021weakly, li2022scale}. However, MIL struggles with variable anomaly durations and is impractical for real-time scenarios due to its reliance on long-term dependencies, which our work seeks to overcome its limitations.

While VAD has been widely studied, VAR is less explored due to the scarcity of labeled anomalies~\cite{sultani2018real}. Recent efforts to apply Large Language and Vision (LLV) models~\cite{radford2021learning, singh2022flava} in VAR have shown promising results~\cite{zanella2023delving, zara2023autolabel}. These models, trained on large image-caption datasets, can improve anomaly recognition. Our work adapts LLV models for weakly supervised VAR, addressing key challenges in this domain.

\subsection{Graph Neural Networks and Knowledge Graphs}

Graph Neural Networks, such as Graph Convolution Networks (GCNs)~\cite{zhang2019graph} and Graph Attention Networks (GATs)~\cite{velivckovic2017graph}, are widely used to process structural data represented as nodes and edges. The Message Passing Network framework~\cite{gilmer2017neural} provides a generalized GNN model, where messages are computed for all edges, aggregated by nodes, and updated to generate node embeddings. These embeddings capture the graph’s structural essence and are critical for tasks requiring an understanding of relationships between entities.

GNNs are often combined with knowledge graphs to enhance tasks like question answering~\cite{yasunaga2021qa}, information retrieval~\cite{gao2022graph}, and reasoning~\cite{gao2020multi}. A key resource in this space is ConceptNet~\cite{speer2017conceptnet}, a large-scale semantic knowledge graph that encodes relationships between words, such as ``IsA" or ``RelatedTo," across multiple languages. ConceptNet is widely used in various domains to provide structured context and reasoning capabilities. In our work, ConceptNet serves as one of the core components of our automated knowledge graph generation framework, supplying structured knowledge that aids in anomaly recognition.

While previous approaches~\cite{zhong2019graph, ullah2023ad} have used GNNs for VAD, they mainly focused on capturing temporal relationships between video frames. In contrast, our method employs a novel GNN architecture over a knowledge graph constructed through automated generation to perform semantic reasoning for VAR tasks, extending GNN capabilities beyond temporal analysis.

\subsection{Large Multimodal Models and Prompt Optimization}

Large Language and Vision models~\cite{radford2021learning, singh2022flava}, which integrate text and image data in a joint embedding space, have proven successful in various tasks, including anomaly detection~\cite{xu2021videoclip, wang2023actionclip}. These models use modality-specific encoders and contrastive learning to align representations across different modalities. AnomalyCLIP~\cite{zanella2023delving}, which leverages CLIP~\cite{radford2021learning}, demonstrated improved VAR performance by harnessing the strengths of LLV models. Expanding on this, recent models like ImageBind~\cite{girdhar2023imagebind} support multiple modalities. Our work builds on this progress by incorporating ImageBind to explore multimodal anomaly detection.

Prompt optimization techniques were introduced to enhance model performance by refining embedding representations. Initial methods relied on manually crafted prompts~\cite{kojima2022large}, followed by learnable prompt techniques~\cite{li2021prefix, lester2021power}, which automatically optimize prompts using gradients. However, these approaches often face overfitting and scalability issues~\cite{zhou2022conditional}. To overcome these limitations, our research replaces prompt optimization with knowledge graph reasoning, which acts as knowledge extraction from a large language model, providing a more efficient and scalable approach for improving model efficacy.

%% file: Sections/3_Methodology.tex
\subsection{Overall Pipeline}

Our proposed approach consists of two major frameworks: mission-specific knowledge graph generation and hierarchical graph neural network-based detection. First, the mission-specific KG generation framework is used for generating KG that can extract useful information from a given frame image. In VAR, each mission-specific KG indicates a KG that contains structured knowledge regarding the corresponding anomaly type. Through this automated framework, our method builds $n$ mission-specific KG each corresponding to one of the $n$ anomaly types. The generated mission-specific KGs are used in the following framework, hierarchical GNN-based detection. $n$ hierarchical GNNs are paired with one of the $n$ mission-specific KGs to conduct reasoning over the extracted knowledge to conduct the corresponding mission which is detecting a certain type of anomaly. And large multimodal model's embedding of each frame is processed by each of the GNN models. Now, embeddings from the final embedding node of each GNN model are concatenated into a single embedding vector that represents reasoning results from all mission-specific KGs. The final vector now is conveyed to a short-term temporal model with the previous consecutive 29 frames. Note that this temporal model only considers short-term relationships which does not require long frame sequence as an input. The resulting output from the short-term temporal model passes through a final decision model to have a final score distribution for each anomaly.

Details are discussed in the following subsections: we first address our mission-specific KG generation framework in~\autoref{sec:kgframework}, then the proposed hierarchical multimodal GNN model and short-term temporal model are introduced in~\autoref{sec:gnn} and \autoref{sec:temporal} respectively, and finally training losses are described in~\autoref{sec:training}.

\subsection{Mission-Specific Knowledge Graph Generation Framework}\label{sec:kgframework}

\begin{figure*}[!t]
    \centering
    \includegraphics[width=1.\textwidth]{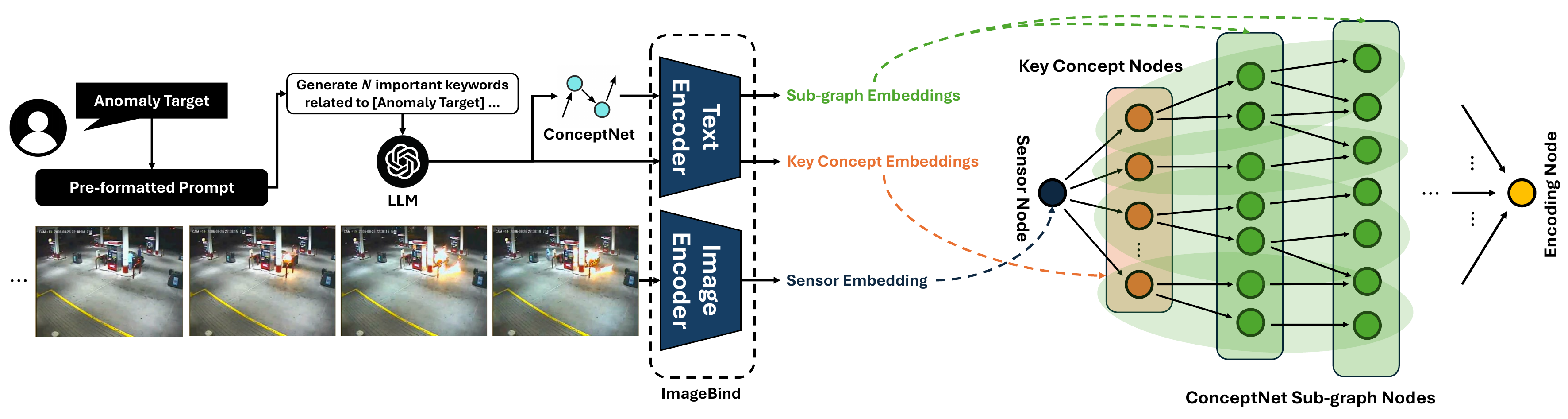}
    \caption{\textbf{The framework for mission-specific knowledge graph generation}.}\label{fig:KGBuild}
\end{figure*}

\autoref{fig:KGBuild} shows the overall framework for mission-specific knowledge graph generation. Given a mission $m_i$ from a user, the framework generates a KG $G_{m_i} = \left(V_{m_i}, E_{m_i}\right)$, which is directed acyclic graph (DAG). In order to generate $G_{m_i}$, it first, combines the given mission $m_i$ with a pre-formatted prompt to have a query message. Then, $n_{cpt}$ number of key concept words $w^{cpt}_{m_i, j}$ are extracted by querying the generated query message to a large language model. Each word $w^{cpt}_{m_i, j}$ is considered as a key concept node $v^{cpt}_{m_i, j}$ and those key concept nodes build part of the KG $\{v^{cpt}_{m_i, j}\} = V^{cpt}_{m_i}\subset V_{m_i}$. Next, to expand KG, iterative sub-graph extraction is conducted using the large language model. First sub-graph nodes $v^{sub_1}_{m_i, j} \in V^{sub_1}_{m_i}$ are extracted by querying the key concept nodes to the large language model. A set of edges containing neighboring relationships $E^{sub_1}_{m_i} = \{\left(v, u\right) \in V^{cpt}_{m_i}\times V^{sub_1}_{m_i} | \left(v, u\right)\in \textbf{ConceptNet} \}$ is also extracted during the process. In our implementation, we only considered the ``\textit{/r/RelatedTo}'' relationship type. Next sub-graph nodes $v^{sub_2}_{m_i, j} \in V^{sub_2}_{m_i}$ are extracted with the similar way but directly neighboring to $V^{sub_1}_{m_i}$ with edges $E^{sub_2}_{m_i} = \{\left(v, u\right) \in V^{sub_1}_{m_i}\times V^{sub_2}_{m_i} | \left(v, u\right)\in \textbf{ConceptNet} \}$. This process, which is extracting $k + 1^{th}$ set of sub-graph nodes $V^{sub_{k + 1}}_{m_i}$ and edges $E^{sub_{k + 1}}_{m_i}$ from $V^{sub_{k}}_{m_i}$, is iteratively conducted until $V^{sub_{d_{sub}}}_{m_i}$ and $E^{sub_{d_{sub}}}_{m_i}$ are retrieved where $d_{sub}$ indicates maximum depth of the sub-graph. The extracted sub-graph nodes $V^{sub}_{m_i} = \bigcup_{1\leq k\leq d_{sub}}{V^{sub_{k}}_{m_i}}$ and edges $E^{sub}_{m_i} = \bigcup_{1\leq k\leq d_{sub}}{E^{sub_{k}}_{m_i}}$ also build part of the KG by $V^{sub}_{m_i} \subset V_{m_i}$ and $E^{sub}_{m_i} \subset E_{m_i}$. Finally, $G_{m_i}$ is completed by combining sensor node $v^{snr}_{m_i}$ and encoding node $v^{ecd}_{m_i}$. $v^{snr}_{m_i}$ is attached to the graph by connecting it with key concept nodes $v^{cpt}_{m_i, j}$ towards $v^{cpt}_{m_i, j}$ direction while $v^{ecd}_{m_i}$ is attached by connecting it with leaf nodes, which are nodes that do not have outgoing edges, of sub-graph nodes towards $v^{ecd}_{m_i}$ direction. Completed $G_{m_i}$ is described in~\autoref{eq:completedkg}.

\begin{align}
\begin{split}\label{eq:completedkg}
  V_{m_i} = & \left\{v^{snr}_{m_i}\right\}\cup V^{cpt}_{m_i}\cup V^{sub}_{m_i}\cup \left\{v^{ecd}_{m_i}\right\} \\
  E_{m_i} = & \left\{\left(v^{snr}_{m_i}, u\right) | u\in V^{cpt}_{m_i}\right\} \\
  & \cup E^{sub}_{m_i} \\
  & \cup \left\{\left(v, v^{ecd}_{m_i}\right) | v\in V^{sub}_{m_i}\wedge \left(v, u\right)\notin E^{sub}_{m_i} \forall u\in V^{sub}_{m_i} \right\}
\end{split}
\end{align}

To process generated KG $G_{m_i}$ using GNN later, it is necessary to have embeddings of the nodes. For having embedding representations, we use text encoder $\mathcal{E}_T(.)$ for embedding key concept nodes and sub-graph nodes and image encoder $\mathcal{E}_I(.)$ for embedding the sensor node. By using encoders $\mathcal{E}_T(.)$ and $\mathcal{E}_I(.)$ from the pre-trained ImageBind model, we can have embeddings in the same embedding space for different modalities. For the encoding node, it is initialized with a $0$ valued vector. In VAR, multiple mission-specific KGs $G_{m_i}$ are generated by setting each mission $m_i$ as detecting each anomaly type $T_i$. As a result, for the VAR task of $n$ number of anomaly types, a total of $n$ mission-specific KGs $G_{T_i}$ are built by the framework.

\subsection{Hierarchical Multimodal GNN}\label{sec:gnn}

\begin{figure*}[!t]
    \centering
    \includegraphics[width=1.\textwidth]{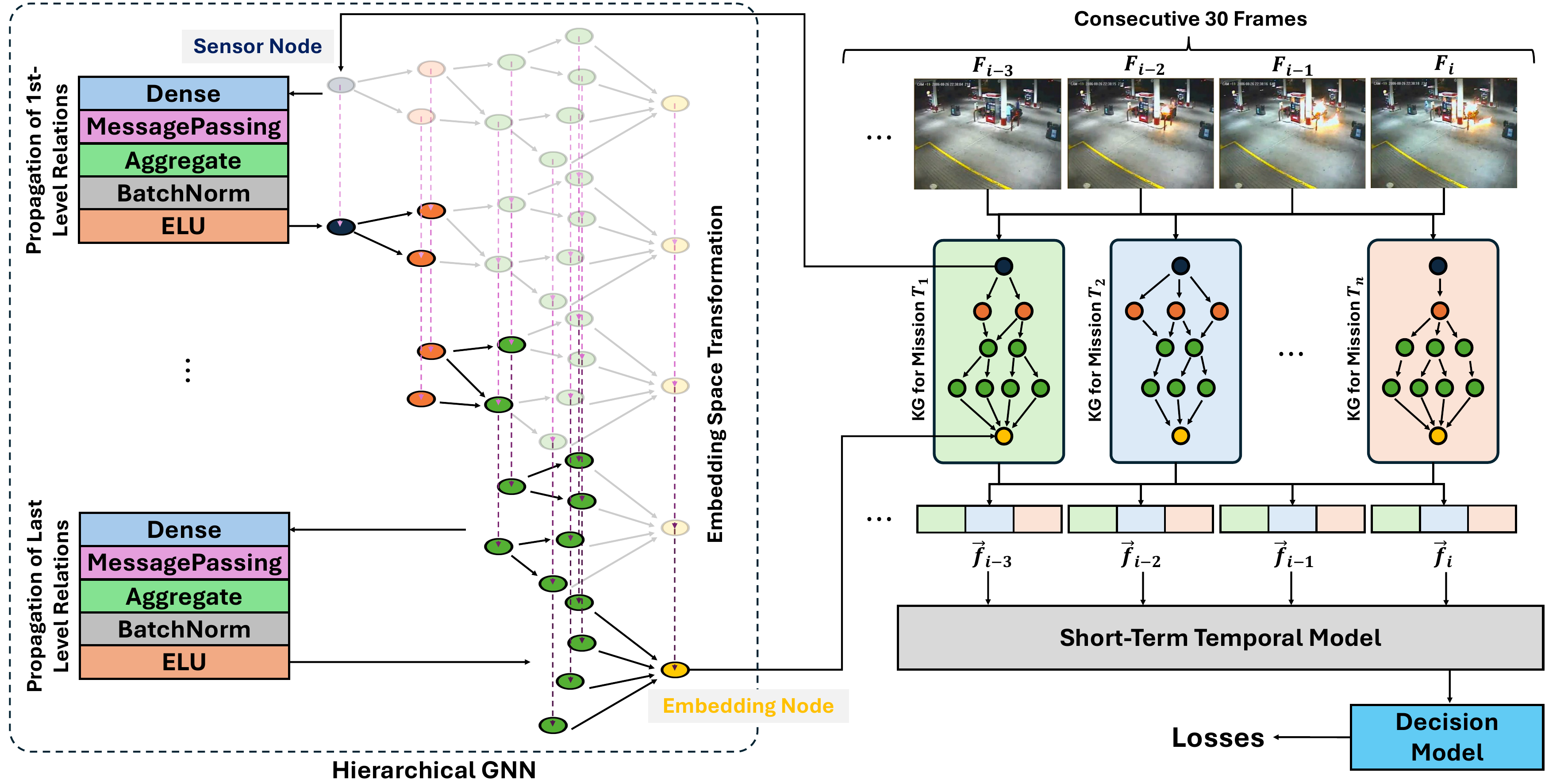}
    \caption{\textbf{The overall framework for our proposed model utilizing the novel concept of hierarchical graph neural network}.}\label{fig:HierarchicalGNN}
\end{figure*}

For a given frame image $F_t$ and mission-specific KGs $G_{T_i}$, to conduct reasoning for each KG, our proposed hierarchical multimodal GNN model is used over each given KG as presented in~\autoref{fig:HierarchicalGNN}. For each KG $G_{T_i}$, the reasoning process propagates from a sensor node of $G_{T_i}$ to its embedding node. To conduct the reasoning process, first, a given frame image $F_t$ is encoded by $\mathcal{E}_I(F_t)$ and used as an embedding vector of the sensor node $v^{snr}_{m_i}$. Next, $d_{sub} + 2$ GNN layers are applied hierarchical way so that reasoning by GNN layers can propagate starting from the sensor node and the final resulting embedding can be located in the embedding node. Each GNN layer $\mathcal{G}_{m_i, l}(.)$ consists of 5 sub-layers: dense layer, hierarchical message passing layer, hierarchical aggregate layer, batch normalization layer, and activation layer. Dense layer $\phi_{m_i, l}\colon \mathbb{R}^{|V_{m_i}|\times D_{l-1}} \to \mathbb{R}^{|V_{m_i}|\times D_{l}}$, where $D_{m_i, l-1}$ indicates dimensionality of previous layer and $D_{m_i, l}$ indicates dimensionality of the current layer, transfers embedding space to have better relationship representations between each knowledge embedding and can be formulated as follows:

\begin{align}
\begin{split}\label{eq:spacetransfer}
  \phi_{m_i, l}(X_{l - 1}) = \mathbf{W}^\phi_{m_i, l}X_{l - 1} + \mathbf{b}^\phi_{m_i, l}
\end{split}
\end{align}

where $\mathbf{W}^\phi_{m_i, l}$ is trainable parametric matrix, $\mathbf{b}^\phi_{m_i, l}$ is trainable bias, and $X_{l-1}\in \mathbb{R}^{|V_{m_i}|\times D_{m_i, l-1}}$ is embeddings of previous layer for nodes. Hierarchical message passing layer $\mathcal{M}^h_{m_i, l}\colon \mathbb{R}^{|V_{m_i}|\times D_{m_i, l}} \to \mathbb{R}^{|E^{(l)}_{m_i}|\times D_{m_i, l}}$ compute messages each corresponding to each node where $E^{(l)}_{m_i} \subset E_{m_i}$ indicates a set of edges that connect depth $l - 1$ nodes $V^{(l-1)}_{m_i} \subset V_{m_i}$ and depth $l$ nodes $V^{(l)}_{m_i} \subset V_{m_i}$. Note that depth $0$ node set $V^{(0)}_{m_i}$ contains only one node which is sensor node $v^{snr}_{m_i}$ and $l$ starts from $1$. Each message indicates measured relativity by the element-wise product between two knowledge embeddings as described below:

\begin{align}
\begin{split}\label{eq:messagepassing}
  \mathcal{M}^h_{m_i, l}(X) = \{X_s\cdot X_d\}_{(s, d)\in E^{(l)}_{m_i}}
\end{split}
\end{align}

where $s$ indicates source node, $d$ indicates destination node, and $X\in \mathbb{R}^{|V_{m_i}|\times D_{m_i, l}}$ is node embeddings. Hierarchical aggregate layer $\mathcal{A}^h_{m_i, l}\colon \{\mathbb{R}^{|V_{m_i}|\times D_{m_i, l}}, \mathbb{R}^{|E^{(l)}_{m_i}|\times D_{m_i, l}}\} \to \mathbb{R}^{|V_{m_i}|\times D_{m_i, l}}$ aggregates all the passed messages from source nodes into a single vector by getting a mean vector meanwhile remaining other embeddings of nodes that do not have messages from their source nodes, denoted as:

\begin{small}
\begin{align}
\begin{split}\label{eq:aggregate}
  &\mathcal{A}^h_{m_i, l}(X, M) =\\ & \left\{X_{d}\mathbf{1}(d\notin V^{(l)}_{m_i}) + \frac{\sum_{(s, d) \in E^{(l)}_{m_i}}{M_{s, d}}}{|\{s \in V_{m_i} | (s, d)\in E^{(l)}_{m_i}\}|}\mathbf{1}(d\in V^{(l)}_{m_i}) \right\}_{d\in V_{m_i}}
\end{split}
\end{align}
\end{small}

where $X\in \mathbb{R}^{|V_{m_i}|\times D_{m_i, l}}$ indicates node embeddings, $M\in \mathbb{R}^{|E^{(l)}_{m_i}|\times D_{m_i, l}}$ indicates messages from $\mathcal{A}^h_{m_i, l}$, and $\mathbf{1}(.)$ indicates a function that gives $1$ if a given statement is true otherwise $0$. The final form of the GNN layer at layer $l$ is formulated as follows:

\begin{align}
\begin{split}\label{eq:gnnlayer}
  &\mathcal{G}_{m_i, l}(X_{l-1}) = \text{ELU}(\\&\text{BatchNorm}\left(\mathcal{A}^h_{m_i, l}\left(\phi_{m_i, l}\left(X_{l - 1}\right), {M}^h_{m_i, l}\left(\phi_{m_i, l}\left(X_{l - 1}\right)\right)\right)\right))
\end{split}
\end{align}

where $\text{ELU}(.)$ indicates Exponential Linear Unit (ELU) activation function. Through this, knowledge embeddings of the current layer are computed $X_l = \mathcal{G}_{m_i, l}(X_{l-1}) \in \mathbb{R}^{|V_{m_i}|\times D_{m_i, l}}$ from previous layer embeddings $X_{l-1} \in \mathbb{R}^{|V_{m_i}|\times D_{m_i, l-1}}$. It is notable that unlike $\mathcal{M}^h_{m_i, l}$ and $\mathcal{A}^h_{m_i, l}$ where it is conducted only for $V^{(l)}_{m_i}$ nodes and $E^{(l)}_{m_i}$ edges, other components are applied for all nodes $V_{m_i}$ and edges $E_{m_i}$ so that embeddings of every node can be placed in the same embedding space for the future reasoning propagation. The final embedding of the KG reasoning $\vec{r}_{T_i}$ is extracted from $v^{ecd}_{m_i}$ of $X_{d_{sub}+2}$ which is the final embedding layer of KG $G_{T_i}$. Finally, reasoning embedding $\vec{f}_t$ of the frame image encoding $\mathcal{E}_I(F_t)$ is formed by concatenating all $\vec{r}_{T_i}$ from each KG as $\vec{f}_t = \vec{r}_{T_1}\frown\vec{r}_{T_2}\frown\cdots\frown\vec{r}_{T_n} \in \mathbb{R}^{D}$ where $D = \sum_{i}{D_{m_i, d_{sub}+2}}$.

\subsection{Short-term Temporal Model}\label{sec:temporal}

Using multiple KGs and reasoning over them with our proposed hierarchical GNNs only considers information from a single frame. To extract reacher information by exploiting video data, we use transformer-based temporal model $\mathcal{T}\colon \mathbb{R}^{T\times D} \to \mathbb{R}^{D}$ where $T$ indicates the number of consecutive frames. Specifically, we use simple a single transformer encoder layer~\cite{vaswani2017attention} and put a sequence of reasoning embeddings including reasoning embeddings of previous $T-1$ consecutive frames $\vec{F}_t = \{\vec{f}_{t - T + 1}, \vec{f}_{t - T + 2}, \cdots, \vec{f}_t\}\in\mathbb{R}^{T\times D}$ to the transformer and only takes the last output embedding $\vec{f}'_t = \mathcal{T}(\vec{F}_t) \in\mathbb{R}^D$ which is corresponding to the last input $\vec{f}_t$. This process is also presented in~\autoref{fig:HierarchicalGNN}. Unlike previous methods that considered not only short-term temporal relationships but also long-term temporal relationships, our model focuses on short-term relationships only by setting $T = 30$ without any video segmentation. 
In VAR, we assume that anomalies occur in short time intervals by following human intuition, i.e. when people encounter anomalies such as gunshots, they can detect them immediately without having to understand long-term relationships.
% CHANGED 
% We assume that in VAR, since anomalies happen in a short time window and when human faces anomaly such as gunshot, they can detect it at once, capturing long-term relationships is not necessary. 
The output of the short-term temporal model $\vec{f}'_t$ is then processed by a decision model $f^{dec}\colon \mathbb{R}^D \to \mathbb{R}^{n + 1}$ which is consists of a simple linear layer and softmax function:

\begin{align}
\begin{split}\label{eq:decisionfunc}
  f^{dec}(\vec{f}'_t) = \text{Softmax}(\mathbf{W}^{dec}\vec{f}'_t + \mathbf{b}^{dec})
\end{split}
\end{align}

where $\mathbf{W}^{dec}$ is trainable parametric matrix, and $\mathbf{b}^{dec}$ is trainable bias. The output of the decision model $\vec{s}_t = f^{dec}(\vec{f}'_t)$ contains the estimated probability of the given frame $F_t$ is normal $p_N(F_t) = \vec{s}_{t, 0}$ and probabilities of the frame belong to each anomaly type $i \in \{1, 2, \cdots, n\}$: $p_{A, i}(F_t) = p_{A}(F_t)p_{i|A}(F_t) = \vec{s}_{t, i}$ where $p_A(F_t) = 1 - p_N(F_t)$ indicates probability of abnormal and $p_{i|A}(F_t) = \frac{\vec{s}_{t, i}}{1 - p_N(F_t)}$ indicates conditional probability of anomaly type $i$ when it is anomaly.

\subsection{Weakly-supervised Frame-level Training}\label{sec:training}

\noindent\textbf{Decaying Threshold-based Anomaly Localization.} 
% It is required to estimate normal frames and anomaly frames among frames in the same video that a single type of anomaly is labeled in weakly-supervised learning to have a better performing model.
A better performing model requires estimating normal frames and anomaly frames among frames in the same video, where a single type of anomaly is labeled in weakly-supervised learning.
However, it is difficult to implement a widely used rank-based approach without using MIL-based training. To resolve this issue and have a model that is capable of fully frame-level training and inference, we introduce a novel anomaly localization method, which is a method that separates normal frames from anomaly-labeled video, using our proposed concept of decaying threshold. Decaying threshold $\theta_d$ is denoted as follows:

\begin{align}
\begin{split}\label{eq:decayingth}
  \theta^{(iter)}_d = \frac{\theta_0\alpha^{iter}_d}{2} + 0.5
\end{split}
\end{align}

where $\theta_0$ indicates initial value of $\theta_d$, $\alpha_d$ indicates threshold decay rate, and $iter$ indicates current iteration of training. For each training iteration $iter$, frames $F_t$ from anomaly-labeled videos are considered as normal frames when $p_A(F_t) = 1 - p_N(F_t) \leq 1 - \theta^{(iter)}_d$ is satisfied. In our implementation, we set $\theta_0 = 1$ so that the model initially considers every frame in anomaly-labeled videos as an anomaly frame and slowly approaches $0.5$ as it separates normal frames. We assume that while all reasoning embeddings of frames $F^{A}_t$ from anomaly-labeled videos separated from reasoning embeddings of frames $F^{N}_t$ from normal-labeled videos, the ones $F^{A+}_t$ that are similar to normal-labeled frames ($p_A(F^{A+}_t) \leq 1-\theta^{(iter)}_d$) slowly separated from the anomaly distribution, which makes them be separated later from anomaly distribution $F^{A-}_t$ that separated faster ($p_A(F^{A-}_t) > 1-\theta^{(iter)}_d$) by the decaying threshold.

\noindent\textbf{Cross-entropy Loss.} After the decaying threshold-based anomaly localization, we have normal frames $F^N_t$, normal estimated frames $F^{A+}_t$, and anomaly estimated frames $F^{A-}_t$. We use the cross-entropy loss to maximize probabilities that match each case. Thus, we have three cross-entropy terms for each set of cases as follows:

\begin{align}
  \mathcal{L}_{N} &= -\frac{\lambda_{N}}{|F^N|}\sum_t{\log{p_N(F^N_t)}}\\
  \mathcal{L}_{A+} &= -\frac{\lambda_{N}}{|F^{A+}|}\sum_t{\log{p_N(F^{A+}_t)}}\\
  \mathcal{L}_{A-} &= -\frac{1}{|F^{A-}|}\sum_{t}{\lambda_{A, y_t}\log{p_{A,y_t}(F^{A-}_t)}}
\end{align}

where $y_t$ indicates the corresponding anomaly type, $\lambda_N$ indicates loss weight for normal cross-entropy, and $\lambda_{A, c}$ indicates loss weight for anomaly class $c$ cross-entropy. $\lambda_N$ and $\lambda_{A, c}$ are adjusted to cope with data imbalance.

\noindent\textbf{Sparsity and Smoothing Losses.} We adapted the same two regularization loss terms that are used in~\cite{Sultani_2018_CVPR}. First regularization is done through sparsity loss term exploiting that the anomaly cases are rare in the entire time frames. Sparsity loss tries to minimize the number of frames that are predicted as anomalies by following:

\begin{align}
  \mathcal{L}_{spa} = \frac{1}{|F^{A}|}\sum_{t}{p_A(F^{A}_{t})}
\end{align}

Note that it is only applied to frames from anomaly-labeled videos. Also, to make predictions smoother along the time dimension, another loss term smoothing loss is implemented as follows:

\begin{align}
  \mathcal{L}_{smt} = \frac{1}{|F|}\sum_{t}{\left(p_A(F_t) - p_A(F_{t-1})\right)^2}
\end{align}

where $F_t$ and $F_{t-1}$ are consecutive two frames.

By combining all loss terms, final loss $\mathcal{L}$ is formed as follows:

\begin{align}
  \mathcal{L} = \mathcal{L}_{N} + \mathcal{L}_{A+} + \mathcal{L}_{A-} + \lambda_{spa}\mathcal{L}_{spa} + \lambda_{smt}\mathcal{L}_{smt}
\end{align}

where $\lambda_{spa}$ and $\lambda_{smt}$ indicate balance coefficients for sparsity loss and smoothing loss term respectively.

%% file: Sections/4_Experiments.tex
\begin{table}
    \centering
    \caption{Memory requirements for computational graph over each model in MiB.}
    \label{tab:memory}
    \begin{adjustbox}{width=0.4\textwidth}
    \begin{tabular}{cccc}
    \toprule
    \multirow{2}{*}{\textbf{Batch Size}} & \multicolumn{2}{c}{\textbf{Model}} & \multirow{2}{*}{\textbf{Efficiency}}  \\
    \cmidrule(lr){2-3}
     & ImageBind & \MissionGNN \\
    \midrule
    1 & 2,171 & 787 & $2.75\times$ \\
    32 & 5,240 & 960 & $5.45\times$ \\
    64 & 5,594 & 1,508 & $3.70\times$ \\
    128 & 5,948 & 2,228 & $2.66\times$ \\
    \bottomrule
    \end{tabular}
  \end{adjustbox}
\end{table}
\subsection{Experimental Settings}

\subsubsection{Implementation Details.} For our multimodal embedding model, we use the pre-trained ImageBind huge model. Also, for the mission-specific KG generation, we use GPT-4 and ConceptNet 5. The loss term balance coefficients $\lambda_{spa}$ and $\lambda_{smt}$ are set to $0.001$ for both of them. We utilize the AdamW~\cite{loshchilov2017decoupled} as the optimizer with learning rate of $10^{-5}$, a weight decay of $1.0$, $\beta_1 = 0.9$, $\beta_2 = 0.999$, and $\epsilon = 10^{-8}$. The decaying threshold $\alpha_d$ is set to $0.9999$. For our GNN models, we set the dimensionality all the same $D_{m_i, l} = 8$. In the case of the short-term temporal model, we use an inner dimensionality of 128 with 8 attention heads. Finally, the model is trained for 3,000 steps with a mini-batch of 128.

\subsubsection{Evaluation Metrics.} We perform evaluations that include both VAD and VAR. Consistent with previous research, VAD performance is measured using the area under the curve (AUC) of the receiver operating characteristics (ROC) at the frame level.
A higher AUC at the frame level indicates a better ability to distinguish normal from abnormal events. For the VAR evaluation, we extend the AUC metric to the multi-classification scenario. Specifically, we compute the AUC for each anomalous class, treating anomalous frames of the class as positive instances and all others as negative instances. We then compute the mean AUC (mAUC) across all anomalous classes. Likewise, for the XD-Violence dataset, we follow the established evaluation protocol~\cite{Wu2020not} and VAD results are presented using the average precision (AP) derived from the precision-recall curve (PRC). For VAR outcomes, we report the mean average precision (mAP), which involves averaging the binary AP values across all abnormal classes.

\begin{table*}
    \centering
    \caption{Results of the state-of-the-art methods and \MissionGNN on UCF-Crime}
    \label{tab:ucf_crime}
    \begin{adjustbox}{width=1.\textwidth}
    \begin{tabular}{cccccccccccccccc}
    \toprule
    \multirow{2}{*}{\textbf{Method}} & \multicolumn{13}{c}{\textbf{Class}} & \multirow{2}{*}{\textbf{mAUC}} & \multirow{2}{*}{\textbf{AUC}}  \\
    \cmidrule(lr){2-14}
     & Abuse & Arrest & Arson & Assault & Burglary & Explosion & Fighting & RoadAcc. & Robbery & Shooting & Shoplifting & Stealing & Vandalism &\\
    \midrule
    
    CLIP zero-shot~\cite{radford2021learning} & 57.37 & 80.65 & 91.72 & 80.83 & 74.34 & 90.31 & 83.54 & 87.46 & 70.22 & 63.99 & 71.21 & 45.49 & 66.45 & 74.12 & 58.63 \\
    
    \midrule
    RTFM~\cite{tian2021weakly}  & 79.99 & 62.57 & 90.53 & 82.27 & 85.53 & 92.76 & 85.21 & 90.31 & 81.17 & 82.82 & 92.56 & 90.23 & 87.20 & 84.86 & 84.03 \\
    
    S3R~\cite{wu2022self}  & 86.38 & 68.45 & 92.19 & 93.55 & 86.91 & 93.55 & 81.69 & 85.03 & 82.07 & 85.32 & 91.64 & 94.59 & 83.82 & 86.55 & 85.99\\
    
    SSRL~\cite{li2022scale}  & \textbf{95.33} & 79.26 & 93.27 & 91.74 & 89.06 & 92.25 & 87.36 & 80.24 & \textbf{87.75} & 84.5  & 92.31 & 94.22 & 88.17 & 88.88 & \textbf{87.43} \\
        
    ActionCLIP~\cite{wang2021actionclip} & 91.88 & 90.47 & 89.21 & 86.87 & 81.31 & 94.08 & 83.23 & 94.34 & 82.82 & 70.53 & 91.60  & 94.06 & \textbf{89.89} & 87.71 & 82.30 \\
    
    AnomalyCLIP~\cite{zanella2023delving} & 75.03 & \textbf{94.56} & \textbf{96.66} & 94.80  & 90.08 & \textbf{94.79} & 88.76 & 93.30  & 86.85 & 87.45 & 89.47 & 97.00 & 89.78 & 90.66 & 86.36 \\
    
    \midrule

    \makecell{AnomalyCLIP~\cite{zanella2023delving} \\ w/ Frame Restriction} & \makecell{69.51 \\ \textcolor{red}{(-5.52)}} & \makecell{90.03 \\ \textcolor{red}{(-4.53)}} & \makecell{92.44 \\ \textcolor{red}{(-4.22)}} & \makecell{88.47 \\ \textcolor{red}{(-6.33)}} & \makecell{82.38 \\ \textcolor{red}{(-7.7)}} & \makecell{90.44 \\ \textcolor{red}{(-4.35)}} & \makecell{85.83 \\ \textcolor{red}{(-2.93)}} & \makecell{90.64 \\ \textcolor{red}{(-2.66)}}  & \makecell{82.22 \\ \textcolor{red}{(-4.63)}} & \makecell{83.21 \\ \textcolor{red}{(-4.24)}} & \makecell{87.43 \\ \textcolor{red}{(-2.04)}} & \makecell{93.95 \\ \textcolor{red}{(-3.05)}} & \makecell{83.48 \\ \textcolor{red}{(-6.3)}} & \makecell{86.15 \\ \textcolor{red}{(-4.5)}} & \makecell{83.35 \\ \textcolor{red}{(-3.01)}} \\
    
    \midrule
    
    \textbf{\MissionGNN (Ours)} & 93.68 & 92.95 & 94.15 & \textbf{96.89}  & \textbf{92.31} & \textbf{94.79} & \textbf{92.38} & \textbf{97.16}  & 81.40 & \textbf{90.57} & \textbf{92.94} & \textbf{97.74} & 84.03 & \textbf{92.38} & 84.48 \\
    \bottomrule
    \end{tabular}
  \end{adjustbox}
\end{table*}

\begin{table*}
    \centering
    \caption{Results of the state-of-the-art methods and \MissionGNN on XD-Violence}
    \label{tab:xd_violence}
    \begin{adjustbox}{width=0.6\textwidth}
    \begin{tabular}{ccccccccc}
    \toprule
    \multirow{2}{*}{\textbf{Method}} & \multicolumn{6}{c}{\textbf{Class}} & \multirow{2}{*}{\textbf{mAP}}&\multirow{2}{*}{\textbf{AP}}  \\
    \cmidrule(lr){2-7}
     & Abuse & CarAccident & Explosion & Fighting & Riot & Shooting & \\
    \midrule
    CLIP zero-shot~\cite{radford2021learning} & 0.32 & 12.21 & 22.26 & 25.25 & 66.60 & 1.26 & 21.32 & 27.21 \\
    \midrule
    RTFM~\cite{tian2021weakly}  & 9.25 &  25.36 & 53.53 & 61.73 & 90.38 & 18.01 & 43.04 & 77.81 \\

    S3R~\cite{wu2022self} & 2.63 & 23.82 & 45.29 & 49.88 & 90.41 & 4.34 & 36.06 & 80.26 \\
    
    ActionCLIP~\cite{wang2021actionclip} & 2.73 & 25.15 & 55.28 & 58.09 & 87.31 & 12.87 & 40.24 & 61.01 \\
    
    AnomalyCLIP~\cite{zanella2023delving} & 6.10 & \textbf{31.31} & \textbf{68.75} & \textbf{71.44}  & \textbf{92.74} & \textbf{26.13} & \textbf{49.41} & 78.51 \\
    
    \midrule
    
    \makecell{AnomalyCLIP~\cite{zanella2023delving} \\ w/ Frame Restriction} & \makecell{4.2 \\ \textcolor{red}{(-1.9)}} & \makecell{21.24 \\ \textcolor{red}{(-10.07)}} & \makecell{48.34 \\ \textcolor{red}{(-20.41)}} & \makecell{47.08 \\ \textcolor{red}{(-24.36)}}  & \makecell{73.33 \\ \textcolor{red}{(-19.41)}} & \makecell{16.53 \\ \textcolor{red}{(-9.6)}} & \makecell{35.12 \\ \textcolor{red}{(-14.29)}} & \makecell{65.51 \\ \textcolor{red}{(-13)}} \\
    
    \midrule
    \textbf{\MissionGNN (Ours)} & \textbf{10.98} & 8.91 & 42.49 & 34.06 & 81.64  & 15.03  & 32.18 & \textbf{98.42} \\
    \bottomrule
    \end{tabular}
  \end{adjustbox}
\end{table*}

\subsection{Datasets}

Our experiments utilize two Video Anomaly Detection (VAD) datasets: UCF-Crime~\cite{Sultani_2018_CVPR} and XD-Violence~\cite{Wu2020not}. Additionally, the ShanghaiTech dataset~\cite{liu2018ano_pred}, comprising 437 videos captured from diverse surveillance cameras on a university campus, is commonly employed in this research field. However, since some anomaly class labels are only present in the training dataset, we have determined that ShanghaiTech is not suitable for evaluating our method. Therefore, we have chosen not to include it in our evaluation process.
\\
\textbf{UCF-Crime} is an extensive dataset comprising 1900 untrimmed surveillance videos depicting 13 real-world anomalies with significant implications for public safety. The training set comprises 800 normal and 810 anomalous videos, while the testing set includes the remaining 150 normal and 140 anomalous videos. 
\\
\textbf{XD-Violence}, a large-scale violence detection dataset, which consists of 4754 untrimmed videos with audio signals and weak labels. It is partitioned into a training set of 3954 videos and a test set of 800 videos, totaling 217 hours in duration and covering various scenarios with 6 anomaly categories. Notably, each violent video may carry multiple labels, ranging from 1 to 3. To align with our training setup, where only one anomaly type per video is considered, we curate a subset comprising 4463 videos containing at most one anomaly. 

\subsection{Evaluation on Training Memory Efficiency}

To demonstrate the advantage in training in terms of memory consumption of our proposed approach, we measured GPU memory consumption specifically the amount of memory for computing computational graphs which is required to conduct gradient backpropagation. \autoref{tab:memory} shows measured GPU memory consumption used for computational graphs for each of ImageBind and our GNN model by different batch sizes. We can observe that our model consumes $2.66\times $ to $5.45\times $ lower memory. Considering previous approaches used learnable prompts or other types of gradient computation of the LLVs are also using additional models to process the output from the LLV, the memory efficiency can be increased. It enables memory-efficient training while maintaining or outperforming previous models in some benchmarks that are shown in \autoref{sec:VADVAReval}.

\subsection{Evaluation on VAD and VAR}\label{sec:VADVAReval}
\autoref{tab:ucf_crime} and \autoref{tab:xd_violence} show the evaluation results of both VAD and VAR tasks on UCF-Crime dataset and XD-Violence dataset respectively.
\MissionGNN surpasses the performance of cutting-edge VAR tasks in the UCF-Crime dataset, albeit showing a slight lag in VAD tasks. In contrast, AnomalyCLIP exhibits superior performance in VAR tasks within the XD-Violence dataset, while our framework notably outperforms state-of-the-art VAD tasks in the same dataset.

\subsection{Evaluation on Real-time VAD and VAR}

A key advantage of the \MissionGNN framework is its ability to operate with only 30 consecutive frames during both training and inference, making it well-suited for real-time anomaly detection and recognition. In contrast, AnomalyCLIP, which achieves higher performance in some evaluations, requires the entire video sequence for both training and inference, limiting its practical use in real-time applications such as intelligent surveillance.
To illustrate this distinction, we conducted a comparative evaluation on the UCF-Crime and XD-Violence datasets, constraining AnomalyCLIP to the same 30-frame input as our model. The results, shown in \autoref{tab:ucf_crime} and \autoref{tab:xd_violence}, highlight the impact of this limitation on AnomalyCLIP’s performance. In the VAR task, restricting the frame count caused a significant performance drop across all anomaly classes for AnomalyCLIP, resulting in reductions of 4.5\% in mAUC and 14.29\% in mAP scores across both datasets.
This adjustment altered the performance comparison between the two methods. On the UCF-Crime dataset, \MissionGNN exhibited a 6.23\% improvement over AnomalyCLIP, further widening the performance gap. On the XD-Violence dataset, however, the gap narrowed by 3.71\%, reducing the original 18\% difference, which emphasizes the limitations of previous approaches in real-time settings.
In the VAD task, \MissionGNN outperformed AnomalyCLIP on both datasets, with gains of 1.13\% in AUC and 32.91\% in AP scores. These findings underscore the performance degradation of AnomalyCLIP when adapted for real-time analysis, as reflected in its lower AUC and AP scores on both datasets, reinforcing the practical advantages of our proposed framework for real-time anomaly detection and recognition.

%% file: Sections/5_Discussion.tex
\noindent\textbf{Significant Performance Drop of the Previous State-of-the-Art Model on XD-Violence in Real-Time Scenarios.} The notable performance drop of AnomalyCLIP on the XD-Violence dataset in real-time scenarios stems from the dataset’s highly diverse and dynamic nature. XD-Violence features a wide range of video types, including fast-paced scenes from movies and sports events, where frequent changes in camera angles and scene transitions are common. These scene swaps make it unrealistic for real-time applications, as relevant frames may be separated by long periods within the video sequence. AnomalyCLIP performs well in this dataset because it processes entire video sequences, allowing it to consider both past and future frames, effectively capturing long-term relationships and subtle anomalies spread across disparate time segments. This capability is particularly advantageous in offline settings, where the model can access all frames at once, contributing to its superior performance in VAR tasks.
However, in real-time applications, where future frames are not available, this reliance on long-term temporal context becomes a major limitation. When AnomalyCLIP is constrained to operate within a 30-frame window, as required for real-time analysis, it experiences a significant performance drop, since it can no longer rely on distant future or past frames to refine its predictions.
In contrast, \MissionGNN is designed specifically for real-time scenarios, where timely and efficient anomaly detection is crucial. By focusing on short-term relationships within the immediate frame window, \MissionGNN is more suitable for real-world applications like violence alerting, where multiple scenes and unpredictable events make it impractical to rely on future information. Although \MissionGNN may not capture long-term dependencies as effectively, its real-time responsiveness makes it more practical in scenarios requiring immediate detection. This comparison highlights the trade-off between models like AnomalyCLIP, optimized for comprehensive analysis with long-term context, and those like \MissionGNN, which are tailored for efficient, real-time anomaly detection despite scene-swapping challenges in datasets like XD-Violence.

\noindent\textbf{Limitations on mission-specific graph generation.} 
Even though we introduce our GNN reasoning approach to deal with the heavy gradient computation from large pre-trained models, the efficacy of our framework heavily relies on the configuration of each mission-specific KG which can limit the performance of the model. Specifically, the construction of mission-specific KGs is facilitated by the output of GPT-4, which denotes the relevant terms associated with each mission. As a result, particularly in the context of VAR tasks, the selection of inappropriate terms by GPT-4 can significantly impair the model's performance even with our reasoning GNN model. In contrast, AnomalyCLIP adopts gradient-based prompt optimization, which is advantageous for more effective related word vector extraction at the cost of necessitating extensive gradient computations. Thus, there exists a trade-off between the computational cost of prompt optimization. Future efforts will focus on mitigating the trade-off relationship by addressing the constraints of our mission-specific graph generation framework. This study will aid in tackling the trend where many LLMs are accessible solely through APIs, hindering direct access to their model weights.

\noindent\textbf{Limitations on anomaly localization.} 
Given that our decaying threshold-based anomaly localization method shown in \autoref{eq:decayingth}. alters the threshold statically, without taking into account the distribution of video frames, it becomes imperative to meticulously select the hyperparameter $\alpha_{d}^{iter}$ to achieve satisfactory test outcomes. In future works, it is essential to devise a mechanism that eliminates the necessity of employing this hyperparameter and instead enables dynamic adjustment of the threshold during model training.

%% file: Sections/6_Conclusions.tex
In conclusion, our study introduces \MissionGNN a novel hierarchical multimodal GNN-based approach for weakly-supervised video anomaly recognition, leveraging an automatically generated mission-specific knowledge graph to efficiently address the challenges of anomaly detection in video data. By circumventing the computational burdens from gradient computation of large multimodal models and eschewing fixed video segmentation, our model enables fully frame-level training and inference, demonstrating state-of-the-art performance on benchmark datasets while having training efficiency in terms of less GPU memory usage. This work not only marks a significant advancement in VAR but also offers a scalable, practical framework applicable to a broad range of real-world scenarios, paving the way for future innovations in intelligent surveillance and beyond.

%% file: Appendices/appendix.tex
\section*{A. Ablation Study on KG Size Variations}

\autoref{tab:ablation_kg_size} presents the results of our ablation study examining the impact of KG size variations on anomaly detection across different anomaly types. We explored two distinct scenarios: one where the KG is expanded by adding more key concept nodes, effectively widening the KG, and another where the KG is deepened through the incorporation of additional subgraphs from ConceptNet.

Our findings suggest that merely increasing the size of the KG can lead to overfitting, resulting in diminished performance. This underscores the importance of developing tailored, mission-specific KGs, as emphasized in our main paper, rather than simply enlarging existing KGs.

\begin{table}[h]
    \centering
    \caption{Ablation study results on different size of KGs on XD-Violence dataset.}
    \label{tab:ablation_kg_size}
    \begin{adjustbox}{width=0.4\textwidth}
    \begin{tabular}{cccc}
    \toprule
    \textbf{Case} & \makecell{\textbf{\# of Key } \\ \textbf{Concept Nodes}} & \makecell{\textbf{Depth of} \\ \textbf{Subgraphs}} & \textbf{AP} \\
    \midrule
    - & 30 & 1 & \textbf{98.42} \\
    Wide KG & 60 & 1 & 97.66 \\
    Deep KG & 30 & 2 & 97.06 \\
    \bottomrule
    \end{tabular}
  \end{adjustbox}
\end{table}

\section*{B. Ablation Study on Decaying Threshold Method}

\autoref{tab:ablation_decayrate} presents the results of our ablation study investigating the impact of various decay rates on the performance of our proposed decaying threshold-based anomaly localization method, specifically within the UCF-Crime dataset context. The findings demonstrate that our decaying threshold approach significantly enhances performance, evidenced by a 2.37\% increase in the AUC score.

Moreover, the study highlights the critical importance of selecting an appropriate decay rate to maximize performance, with optimal results observed for decay rates $\alpha_d \geq 0.9$. This emphasizes the necessity of fine-tuning the decaying rate for achieving high performance in anomaly detection as we discuss in the main paper.

\begin{table}[h]
    \centering
    \caption{Ablation study results on different decay rate of decaying threshold on UCF-Crime dataset.}
    \label{tab:ablation_decayrate}
    \begin{adjustbox}{width=0.3\textwidth}
    \begin{tabular}{cc}
    \toprule
    \textbf{Deacay Rate} & \textbf{AUC} \\
    \midrule
    0.5 & 63.72 \\
    0.6 & 79.23 \\
    0.7 & 77.16 \\
    0.8 & 81.75 \\
    0.9 & 81.07 \\
    0.99 & \textbf{84.48} \\
    0.999 & 81.57 \\
    0.9999 & 83.57 \\
    1.0 (w/o Decaying Threshold) & 82.11 \\
    \bottomrule
    \end{tabular}
  \end{adjustbox}
\end{table}

\section*{C. Ablation Study on Loss Terms}

\autoref{tab:ablation_lossterms} presents an ablation study evaluating the impact of each loss term in our proposed framework. The evaluation was conducted on the UCF-Crime dataset, measuring AUC scores after removing each loss term from the training loop. The results show that using all loss terms together yields the highest performance, with a significant score drop of at least 5.23\% when any individual loss term is omitted. These findings highlight the critical importance of the proposed loss terms in achieving optimal model performance.

\begin{table}[h]
    \centering
    \caption{Ablation study results on loss terms.}
    \label{tab:ablation_lossterms}
    \begin{adjustbox}{width=0.2\textwidth}
    \begin{tabular}{cc}
    \toprule
    \textbf{Case} & \textbf{AUC} \\
    \midrule
    Without $\mathcal{L}_{N}$ & 60.49 \\
    Without $\mathcal{L}_{spa}$ & 79.25 \\
    Without $\mathcal{L}_smt$ & 77.40 \\
    \midrule
    \textbf{Full Model} & \textbf{84.48} \\
    \bottomrule
    \end{tabular}
  \end{adjustbox}
\end{table}

\begin{table*}
    \centering
    \caption{Ablation study results on different pre-trained joint embedding models.}
    \label{tab:ablation_jointembedding}
    \begin{adjustbox}{width=0.8\textwidth}
    \begin{tabular}{cccc}
    \toprule
    \textbf{Case} & \textbf{Number of Supporting Modalities} & \textbf{Joint Space Dimensionality} & \textbf{AUC} \\
    \midrule
    OpenCLIP (ViT-B-32) & 2 & 512 & 84.12 \\
    ImageBind (imagebind\_huge) & 6 & 1024 & \textbf{84.48} \\
    \bottomrule
    \end{tabular}
  \end{adjustbox}
\end{table*}

\begin{table*}
    \centering
    \caption{Ablation study results on a different number of frames.}
    \label{tab:ablation_jointembedding}
    \begin{adjustbox}{width=0.5\textwidth}
    \begin{tabular}{cccc}
    \toprule
    \textbf{Number of Frames} & \textbf{AUC} & \makecell{\textbf{Inference} \\ \textbf{GPU Memory Usage}} & \makecell{\textbf{Inference} \\ \textbf{Runtime}} \\
    \midrule
    10 & 80.78 & \textbf{13,468 MiB} & \textbf{39 ms} \\
    20 & 83.47 & 13,478 MiB & 43 ms \\
    30 & \textbf{84.48} & 13,540 MiB & 46 ms \\
    \bottomrule
    \end{tabular}
  \end{adjustbox}
\end{table*}

\begin{figure}[!t]
    \centering
    \includegraphics[width=0.5\textwidth]{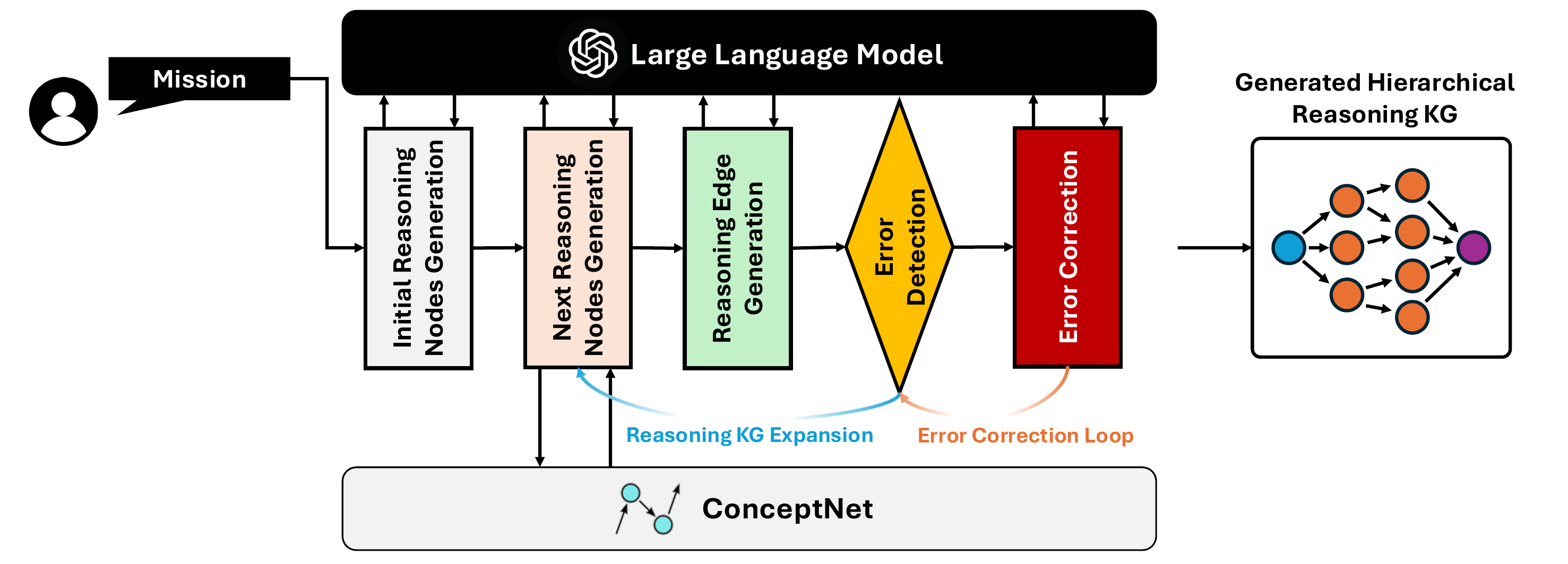}
    \caption{\textbf{Detailed process of mission-specific knowledge graph generation}.}\label{fig:KGBuild}
\end{figure}

\begin{figure*}[!t]
    \centering
    \includegraphics[width=1.0\textwidth]{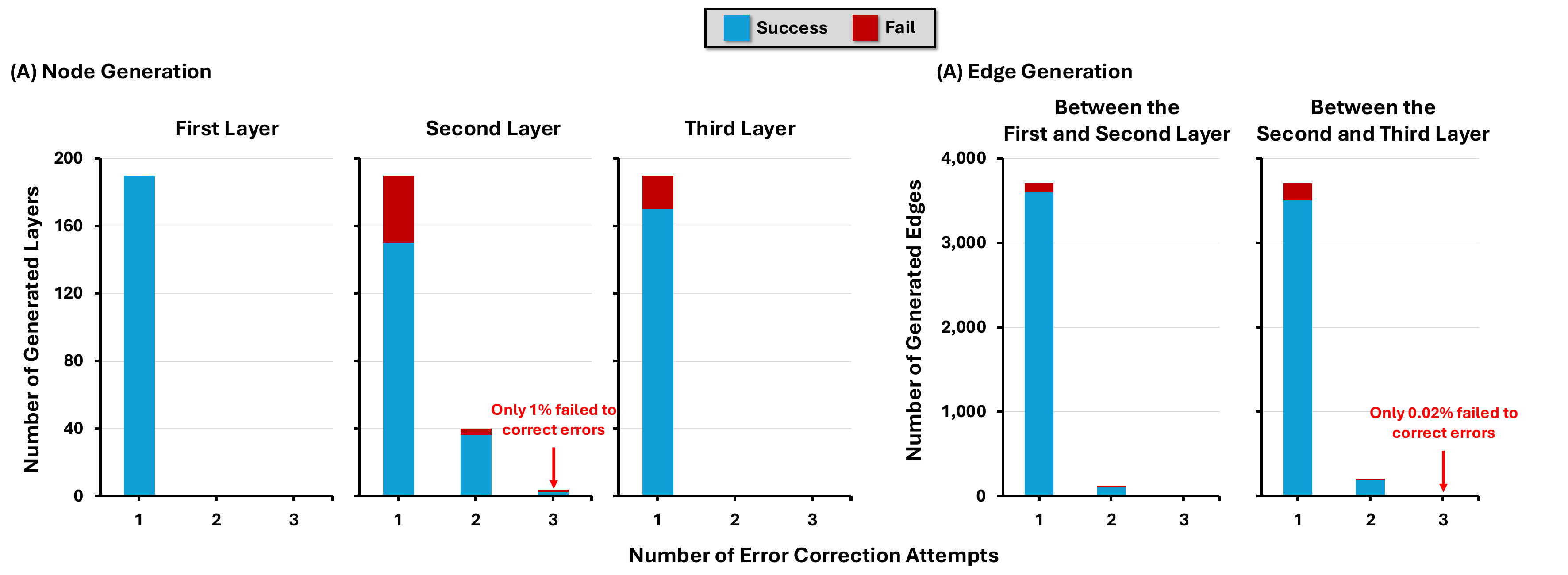}
    \caption{\textbf{Portion of fails on generating valid (A) nodes and (B) edges on each layer by the number of attempts}.}\label{fig:errorcorrection}
\end{figure*}

\begin{figure}[t]
  \centering
    \includegraphics[width=0.5\linewidth]{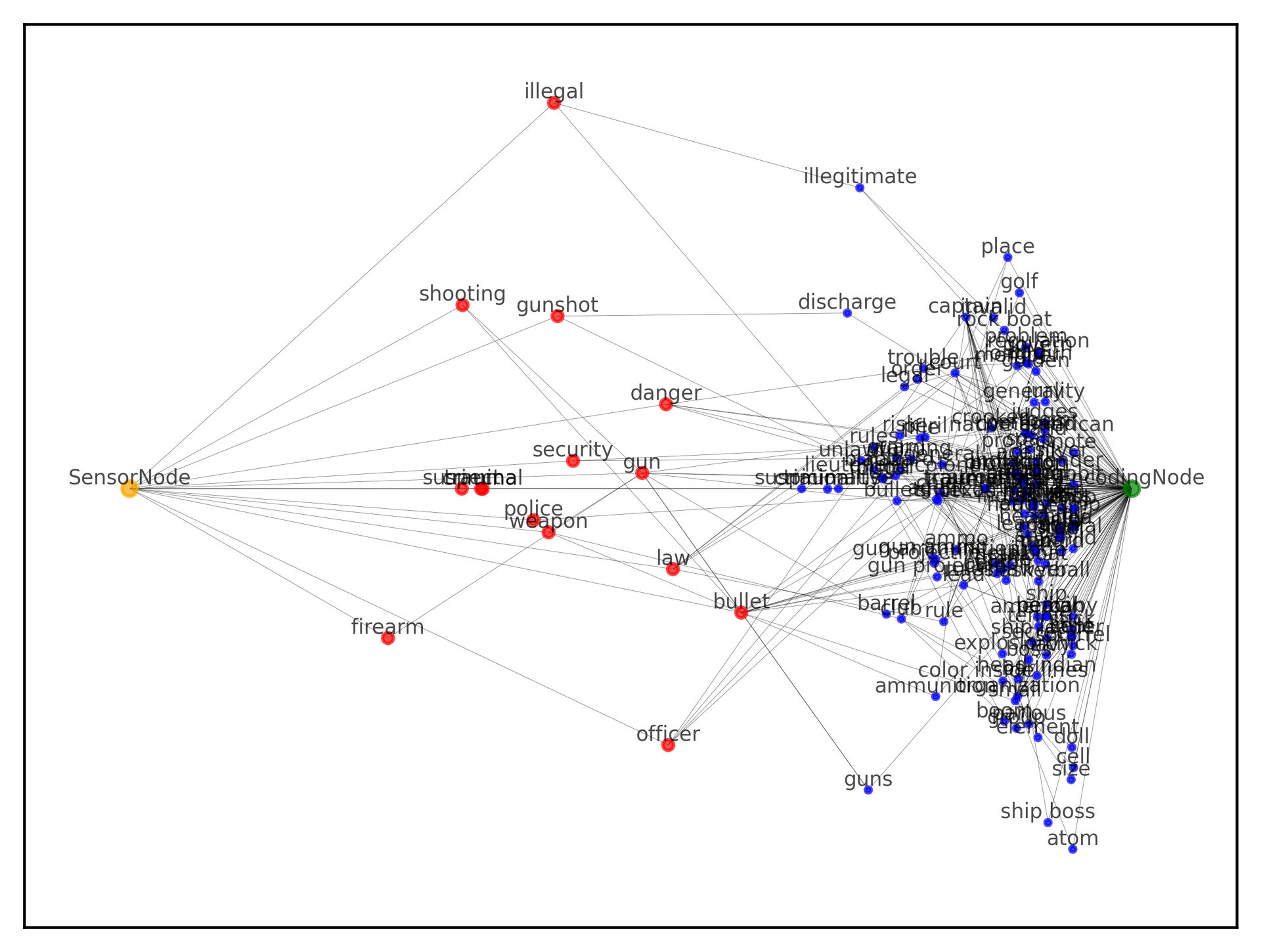}
   \caption{Example of KG for detecting the "Shooting" category in the UCF-Crime dataset. Each color represents: Yellow: Sensor Node, Red: Key Concept Nodes, Blue: Sub-graph Nodes, Green: Encoding Node.}
   \label{fig:KGExample}
\end{figure}

\section*{D. Ablation Study on Joint Embedding Models}

\autoref{tab:ablation_jointembedding} shows the results of our ablation study comparing different joint embedding models. We conducted the evaluation on the UCF-Crime dataset, measuring AUC scores using two joint embedding models: OpenCLIP and ImageBind. Aside from the embedding model and its dimensionality, all other hyperparameters were kept constant. The results demonstrate that our proposed framework performs consistently across both models, with only a 0.36\% difference in scores. This suggests that the reasoning capability of our Hierarchical GNN framework is robust, regardless of the specific joint embedding model used.

\section*{E. Ablation Study on Local Context Size}

\autoref{tab:ablation_jointembedding} presents the results of our ablation study on different local context sizes. Using the UCF-Crime dataset, we measured AUC scores by varying the number of consecutive frames used during inference. The results show a trade-off between performance and resource usage, including GPU memory and runtime. As the local context size increases, the AUC scores improve, but at the cost of slightly higher memory and runtime demands. However, the performance gains tend to level off, with a smaller improvement between 20 and 30 frames compared to the larger gap between 10 and 20 frames. This suggests that after a certain context size, the model’s ability to benefit from additional frames diminishes, likely indicating diminishing returns on performance versus resource consumption.

\section*{F. Knowledge Graph Generation}

\autoref{fig:KGBuild} illustrates the detailed pipeline for mission-specific knowledge graph (KG) generation. The process begins by taking the user’s input and initiating the Initial Reasoning Nodes Generation step. In this phase, initial keywords are generated using a large language model (LLM)—GPT-4 was employed in our evaluation—by utilizing pre-formatted prompts. \autoref{tab:prompts1} provides the prompts used in this step, where SUBJECT represents the user-defined mission, and DUP\_NODES refers to keywords that have appeared in previous layers.

Following the generation of the initial reasoning nodes, the process moves to the Next Reasoning Nodes Generation step, where related keywords are inferred from the previously generated keywords. This step combines the LLM with ConceptNet to extract related keywords. ConceptNet’s output is used to guide the LLM in generating the next set of reasoning nodes. \autoref{tab:prompts2} presents the prompts used in this step, with COMMA\_SEPARATED\_LIST representing the previous layer’s keywords, and SUGGESTED\_KEYWORDS containing related keywords extracted from ConceptNet.

After generating the next set of reasoning nodes, the system identifies logical relations between the nodes from the previous and current layers during the Reasoning Edge Generation step. \autoref{tab:prompts3} shows the prompts for this step, where SUBJECT is replaced with a newly generated keyword, COMMA\_SEPARATED\_LIST contains the keywords from the previous layer, and NOT\_APPEARED\_NODES refers to selected nodes that do not appear in the previous layer, as identified by the LLM.

Once the new layer and corresponding edges have been generated, the framework performs an Error Detection step to validate the correctness of each node and edge according to the definition of our reasoning KG. If any invalid nodes or edges are detected, the Error Correction step is triggered. During this step, errors are corrected using error correction prompts, with a maximum of three correction attempts allowed. If errors persist beyond the third attempt, the erroneous nodes or edges are pruned from the KG. After the error detection and correction loop, the framework returns to the Next Reasoning Nodes Generation step to expand the reasoning KG until the desired number of layers is achieved.

To evaluate the effectiveness of our error correction mechanism, we generated 190 reasoning KGs using our proposed method and tracked error rates by layer and by number of correction attempts, as illustrated in \autoref{fig:errorcorrection}. Across all layers, the number of errors significantly decreases with each correction attempt. For node generation (shown in \autoref{fig:errorcorrection}.(A)), the process results in a maximum of only 1\% pruned nodes by the second layer. Similarly, edge generation (shown in \autoref{fig:errorcorrection}.(B)) results in a maximum of 0.02\% pruned edges. These results demonstrate that the prompts used in our KG generation framework successfully facilitate knowledge extraction from both the LLM and ConceptNet. \autoref{fig:KGExample} presents one KG we generated.

\begin{table*}
    \centering
    \caption{Persona: Initial Reasoning Node Generation}
    \label{tab:prompts1}
    \begin{adjustbox}{width=1.0\textwidth}
    \begin{tabular}{p{2.5cm}c}
    \toprule
    \makecell[c]{\textbf{Type of Prompt}} & \makecell[c]{\textbf{Prompt Format}} \\
    \midrule
    System & 
\makecell[l]{Reference:\\
A knowledge graph have hierarchical levels starting from naive observations to final prediction.\\ 
Each level has the following inference words directly connected only with the previous level words. \\
There are NOT the same words in different levels.\\
\\
Persona:\\
You are a knowledge graph engineer who generates knowledge graph that will help to classify images.\\
\\
Objective:\\
You will be provided a subject.\\
Follow these steps to answer the user queries.\\
\\
Step 1.\\
Observe 20 important words from a image which is related to the provided subject.\\
Do not respond anything for this step.\\
\\
Step 2.\\
Create a comma-separated list of the words that you observed.\\
The comma-separated list you just created is first level of the knowledge graph.\\
Keep in mind.\\
Do not respond anything for this step.\\
\\
Step 3.\\
Print first level of the knowledge graph on the first line.\\
No extraneous text or characters other than the comma-separated list.} \\
    \midrule
    User & 
\makecell[l]{Subject: \{SUBJECT\}}\\
    \midrule
    Error Correction & 
\makecell[l]{The following concepts already appear in previous levels: \{DUP\_NODES\} \\
You must generate new concepts that can be inferred from previous level concepts. \\
Correct this error and give a corrected answer. \\
No extraneous text or characters other than the comma-separated list. \\
Subject: \{SUBJECT\} } \\
    \bottomrule
    \end{tabular}
  \end{adjustbox}
\end{table*}

\begin{table*}
    \centering
    \caption{Persona: Next Reasoning Nodes Generation}
    \label{tab:prompts2}
    \begin{adjustbox}{width=1.0\textwidth}
    \begin{tabular}{p{2.5cm}c}
    \toprule
    \makecell[c]{\textbf{Type of Prompt}} & \makecell[c]{\textbf{Prompt Format}} \\
    \midrule
    System & 
\makecell[l]{Reference:\\
A knowledge graph have hierarchical levels starting from naive observations to final prediction.\\ 
Each level has the following inference words directly connected only with the previous level words. \\
There are NOT the same words in different levels.\\
\\
Persona:\\
You are a knowledge graph engineer who generates knowledge graph that will help to classify images.\\
\\
Objective:\\
You will be provided a subject.\\
And you will be provided comma-separated list which is the previous level of the knowledge graph.\\
And you will be provided suggested keywords.\\
Follow these steps to answer the user queries.\\
\\
Step 1.\\
Create words related to the provided subject which can be explained from combination of several words from previous level.\\
Reference suggested keywords for this step. If you have better keywords, suggest them.\\
Do not respond anything for this step.\\
\\
Step 2.\\
Create a comma-separated list of the words that you just created in step 1.\\
The length of comma-seperated list must be 20.\\
The comma-separated list you just created is next level of the knowledge graph.\\
Keep in mind.\\
Do not respond anything for this step.\\
\\
Print next level of the knowledge graph on the first line.\\
No extraneous text or characters other than the comma-separated list.} \\
    \midrule
    User & 
\makecell[l]{Subject: \{SUBJECT\}\\
Comma-separated list: \{COMMA\_SEPARATED\_LIST\}\\
Suggested keywords: \{SUGGESTED\_KEYWORDS\}}\\
    \midrule
    Error Correction & 
\makecell[l]{The following concepts already appear in previous levels: \{DUP\_NODES\} \\
You must generate new concepts that can be inferred from previous level concepts. \\
Correct this error and give a corrected answer. \\
No extraneous text or characters other than the comma-separated list. \\
Subject: \{SUBJECT\}\\
Comma-separated list: \{COMMA\_SEPARATED\_LIST\}\\
Suggested keywords: \{SUGGESTED\_KEYWORDS\} } \\
    \bottomrule
    \end{tabular}
  \end{adjustbox}
\end{table*}

\begin{table*}
    \centering
    \caption{Persona: Reasoning Edge Generation}
    \label{tab:prompts3}
    \begin{adjustbox}{width=1.0\textwidth}
    \begin{tabular}{p{2.5cm}c}
    \toprule
    \makecell[c]{\textbf{Type of Prompt}} & \makecell[c]{\textbf{Prompt Format}} \\
    \midrule
    System & 
\makecell[l]{Reference:\\
A knowledge graph have hierarchical levels starting from naive observations to final prediction. \\
Each level has the following inference words directly connected only with the previous level words. \\
There are NOT the same words in different levels.\\
\\
Persona:\\
You are a knowledge graph engineer who generates knowledge graph that will help to classify images.\\
\\
Objective:\\
You will be provided a subject and a comma-separated list.\\
Follow these steps to answer the user queries.\\
\\
Step 1.\\
Select maximum 5 words from provided comma-separated list which are related to inferring provided subject.\\
Do not respond anything for this step.\\
\\
Step 2.\\
Create a comma-separated list of the selected words that you observed.\\
Do not respond anything for this step.\\
\\
Step 3.\\
Print the comma-separated list. \\
No extraneous text or characters other than the comma-separated list.} \\
    \midrule
    User & 
\makecell[l]{Subject: \{SUBJECT\}\\
Comma-separated list: \{COMMA\_SEPARATED\_LIST\}}\\
    \midrule
    Error Correction & 
\makecell[l]{The following concepts do not appear in the previous level nodes: \{NOT\_APPEARED\_NODES\} \\
You must select concepts from the previous level concepts that can be important clues to infer the new concept. \\
Correct this error and give a corrected answer. \\
No extraneous text or characters other than the comma-separated list. \\
Subject: \{SUBJECT\}\\
Comma-separated list: \{COMMA\_SEPARATED\_LIST\} } \\
    \bottomrule
    \end{tabular}
  \end{adjustbox}
\end{table*}